\documentclass[letterpaper, 10 pt, conference]{ieeeconf}  

\usepackage{algorithmic}
\usepackage{algorithm}
\usepackage{array}
\usepackage{stfloats}
\usepackage{url}
\usepackage{verbatim}
\IEEEoverridecommandlockouts                              

\usepackage[utf8]{inputenc}                         
\usepackage{graphicx}                               
\usepackage{amsmath}                                
\let\labelindent\relax
\usepackage[inline]{enumitem}                       
\usepackage{textcomp}                               
\usepackage{gensymb}                                
\usepackage{xcolor}                                 
\usepackage{siunitx}                                
\newcolumntype{M}[1]{>{\centering\arraybackslash}m{#1}} 
\usepackage{cite} 
\newcommand{\sgn}{\operatorname{sgn}}

\usepackage{balance} 
\usepackage{makecell}

\usepackage{tabularx}
\newcolumntype{C}[1]{>{\centering\arraybackslash}m{#1}}

\title{\LARGE \bf
Closed-loop Control of Steerable Balloon Endoscopes for Robot-assisted Transcatheter Intracardiac Procedures
}

\author{Max McCandless$^{1\dagger}$, Jonathan Hamid$^{1\dagger}$, Sammy Elmariah$^{2}$, Nathaniel Langer$^{3}$, Pierre E. Dupont$^{1}$
\thanks{$^{1}$Department of Cardiac Surgery, Boston Children's Hospital, Harvard Medical School, Boston, MA, USA.}%
\thanks{$^{2}$Department of Medicine, Cardiology Division, University of California San Francisco, San Francisco, CA, USA.}%
\thanks{$^{3}$Department of Cardiac Surgery, Massachusetts General Hospital, Boston, MA, USA.}%
\thanks{$^{\dagger}$Shared first authorship.}%
\thanks{This work was funded by the National Institutes of Health under grants R01HL167925, R01HL124020, and T32HL007572.}%
\thanks{Max McCandless is the corresponding author (mdm1024@bu.edu).}%
}

\begin{document}

\maketitle
\thispagestyle{empty}
\pagestyle{plain}

\noindent
\begin{abstract}
To move away from open-heart surgery towards safer transcatheter procedures, there is a growing need for improved imaging techniques and robotic solutions to enable simple, accurate tool navigation. 
Common imaging modalities, such as fluoroscopy and ultrasound, have limitations that can be overcome using cardioscopy, i.e., direct optical visualization inside the beating heart. We present a cardioscope designed as a steerable balloon. As a balloon, it can be collapsed to pass through the vasculature and subsequently inflated inside the heart for visualization and tool delivery through an integrated working channel. Through careful design of balloon wall thickness, a single input, balloon inflation pressure, is used to independently control two outputs, balloon diameter (corresponding to field of view diameter) and balloon bending angle (enabling precise working channel positioning). This balloon technology can be tuned to produce cardioscopes designed for a range of intracardiac tasks. To illustrate this approach, a balloon design is presented for the specific task of aortic leaflet laceration. Image-based closed-loop control of bending angle is also demonstrated as a means of enabling stable orientation control during tool insertion and removal. 

\end{abstract}

\section{INTRODUCTION}
\vspace{-1mm}
\noindent
Transcatheter procedures and devices are continuously evolving to overcome the risks and trauma associated with open-heart surgery~\cite{harky_future_2020}. 
However, mastering the complex manipulations needed to control catheters remains difficult. 
In turn, robotic platforms are being developed for improving many existing procedures, such as valve repair and replacement~\cite{nayar_ultrasound-guided_2024,chen_first--human_2024}
, treatment of arrhythmia~\cite{pittiglio_magnetic_2024}
, pacemaker lead placement~\cite{rogatinsky_multifunctional_2023}, and coronary interventions~\cite{li_robust_2025}.  
Robot-assisted steering improves catheter navigation accuracy and enables precise maneuvering to target locations within the beating heart, which has the potential to lower the learning curve for operators and increases opportunities for reduced-risk transcatheter procedures to be performed on more patients~\cite{young_robotic_2022}. 
%

Despite these advancements, both manual and robot-assisted transcatheter procedures are limited by a lack of direct visualization, which inhibits progress and implementation of new technologies, especially those relying heavily on standard imaging modes, such as ultrasound and fluoroscopy.  
Fluoroscopy enables visualization of the catheter, but not the tissue, limiting navigability and increasing the time required to effectively manipulate tools to reach targets, which leads to requiring longer X-ray radiation exposure for patients and clinicians. 
Ultrasound images both the catheter and the tissue, but provides very noisy images, which can obscure anatomical features.

To overcome these imaging limitations, endoscopy inside the blood-filled heart, i.e., cardioscopy, has been developed to provide direct visualization of the contact region between the catheter tip and valve tissue~\cite{vasilyev_three-dimensional_2006, sorensen_camera--tip_2025, rosa_cardioscopically_2017, machaidze_optically-guided_2019, fagogenis_autonomous_2019, padala_transapical_2012, uchida_recent_2011, betensky_use_2012}. 
Cardioscopes are comprised of a clear optical window incorporating a light source and an imaging element. When the window surface is not contacting the anatomy, the blood produces an entirely red image. When pressed against the tissue, however, the window surface displaces the blood and enables direct visualization of the contacted region. 

Cardioscopic imaging has been shown to significantly decrease procedures times and can allow automation of procedures, reducing X-ray exposure times and simplifying procedures~\cite{rosa_cardioscopically_2017, machaidze_optically-guided_2019, fagogenis_autonomous_2019}. 

Prior optical window designs have consisted of either relatively rigid materials of fixed size \cite{vasilyev_three-dimensional_2006, sorensen_camera--tip_2025, rosa_cardioscopically_2017, machaidze_optically-guided_2019, fagogenis_autonomous_2019, padala_transapical_2012} or balloons \cite{uchida_recent_2011, betensky_use_2012}. Since the diameter of the optical window equates to the diameter of the field of view, many fixed size cardioscopes have diameters of $\approx$8-11mm for improved localization capabilities \cite{rosa_cardioscopically_2017, machaidze_optically-guided_2019, fagogenis_autonomous_2019,  padala_transapical_2012}. While devices of this size can be inserted through the heart wall, they are too large for insertion by the preferred and less invasive vascular route.

Optical windows comprised of balloons overcome this limitation. Existing balloon designs, however, possess a single lumen, open at the distal end, which must be shared by the imaging system and any tools \cite{uchida_recent_2011, betensky_use_2012}. Furthermore, since the imaging system is delivered through the working channel, an intermittent \cite{uchida_recent_2011} or continuous \cite{betensky_use_2012} flow of saline must be infused through the channel to clear the visualization area of blood and ensure that blood does not enter the channel. 

Our goal was to create a cardioscope that incorporated the best features of both optical window types – the inflation / deflation capabilities of a balloon and the distinct working channel and imaging compartments of rigid optical windows. Furthermore, to enable precise targeting, we wished to incorporate steerability into the balloon. 

\begin{figure}[t!]
    \centering
    \includegraphics[width=\linewidth]{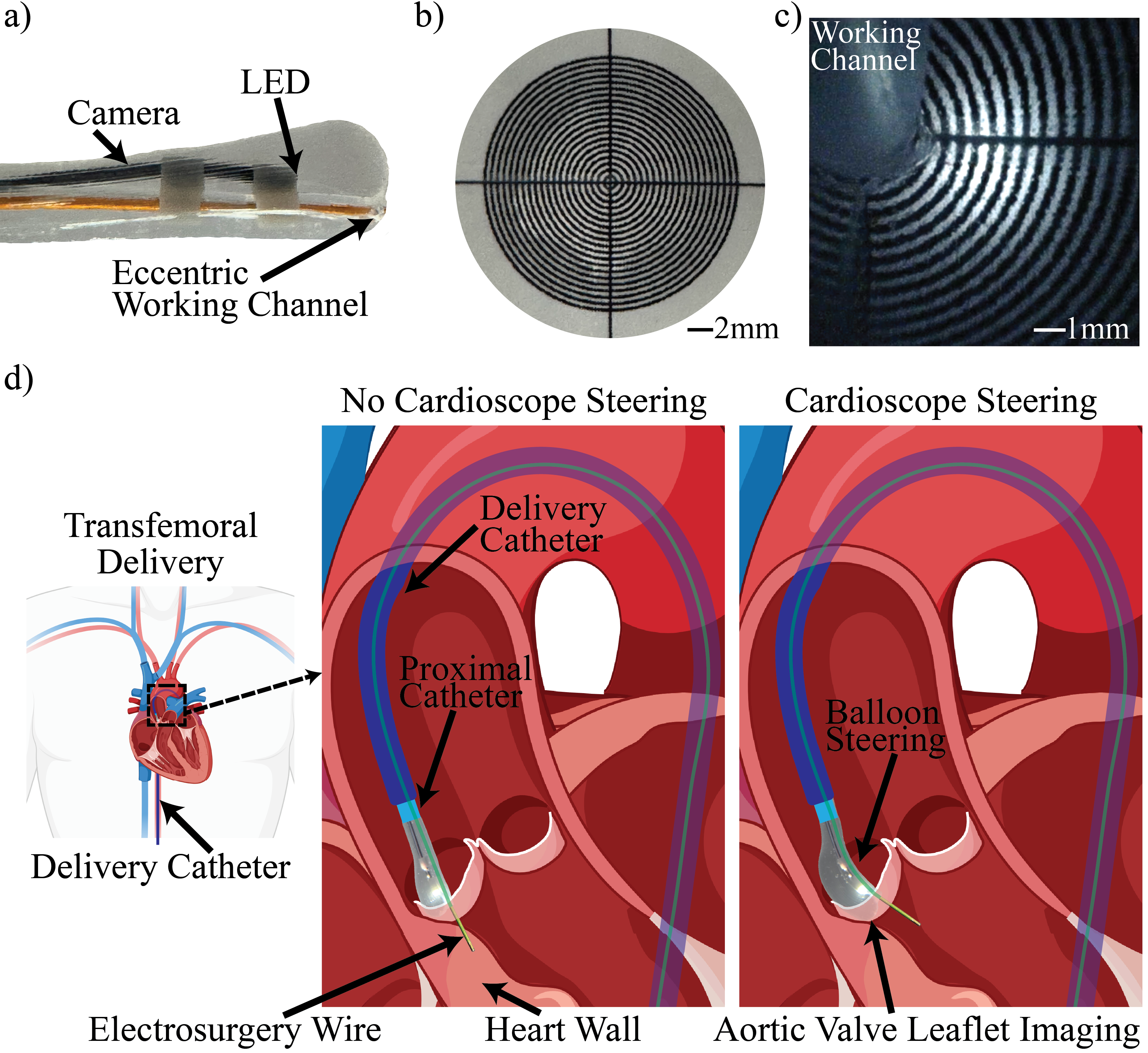}
    \vspace{-7mm}
    \caption{Steerable robotic balloon cardioscope. (a) Balloon cardioscope with camera and LED in balloon interior and separate working channel terminating in the face of the optical window. For this image, in order to see the transparent working channel, an orange tube was placed within it. (b) Target pattern. (c) Target pattern viewed through cardioscope with working channel centered over target. (d) Example clinical application of aortic leaflet laceration. Laceration is initiated by extending an electrosurgical wire through the leaflet. Balloon steerability is used to ensure that the wire is oriented into the ventricle to avoid perforating the heart wall.}
    \vspace{-3mm} 
    \label{FIGURE 1}
\end{figure}

Fig.~\ref{FIGURE 1} illustrates these concepts. 
The balloon is designed such that the imaging system (chip camera and LED light source) lies in its interior while a through-lumen serves as the working channel (Fig.~\ref{FIGURE 1}a). This enables the working channel to be viewed in the image and, consequently, positioned accurately over a target (Fig.~\ref{FIGURE 1}b-c). For the application of aortic leaflet laceration, the value of balloon steerability is shown in Fig.~\ref{FIGURE 1}d. Here, an electrosurgical wire must be passed through the leaflet oriented such that the wire does not accidentally penetrate the heart wall. Balloon steerability avoids the need for an additional concentric steerable sheath.

The paper is arranged as follows. The next section describes cardioscope design and manufacture. In Section~III, the closed-loop controller for balloon inflation and bending is detailed. Section~IV describes validation experiments confirming decoupling of inflation and bending, controller response, and imaging capabilities. Finally, conclusions are presented in Section~V.

\section{Cardioscope Design \& Manufacture}

\noindent
The design and manufacturing process are illustrated here for the clinical application of aortic leaflet laceration. This procedure may be performed just prior to transcatheter aortic valve replacement to prevent the leaflets of the existing valve from occluding the coronary arteries~\cite{khan_transcatheter_2018}. The first step of this procedure uses an electrosurgical wire to create a hole in the center of the base of a leaflet. Cardioscopic visualization enables the operator to ensure they are targeting the correct location and to potentially avoid attempting to traverse calcifications, which are difficult to penetrate. Furthermore, cardioscopic steerability enables the operator to orient the working channel inside the leaflet so that the electrified wire is directed into the ventricle and not out through the heart wall. (Fig. 1d).  

The design specifications for this application are given in Table I. The geometric specifications include balloon inflated and deflated diameters, maximum bending angle and working channel diameter. Steerability requirements specify an angular steering rate, the acceptable amount of steady-state error, and that the optical window face diameter and bending angle be approximately decoupled.  

\begin{table}[tb!]
    \centering
    \caption{Cardioscope specifications for aortic leaflet laceration.}
    \label{Table 1}
    \vspace{-3mm}
    \begin{tabular}{!{\vrule width 1.6pt}C{0.32\linewidth}!{\vrule width 1.6pt}C{0.32\linewidth}!{\vrule width 1.6pt}C{0.185\linewidth}!{\vrule width 1.6pt}}
        \Xhline{1.6pt}
        Procedure Requirements & Steerable Balloon System Defined Metrics & Target Parameters \\
        \Xhline{1.6pt}
        1) Deployable through steerable sheaths  & Maximum collapsed outer diameter ($D_{1}$)  & $D_{1}$ $\leq$ 5mm \\
        \hline
        2) Sufficient viewing area for localization  & Deployed optical face diameter ($D_2$) & 8mm $\leq$ $D_2$ $\leq$ 11mm \\
        \hline
        3) Working channel steerability to targets & Tip deflection angle ($\alpha$)  & $\alpha_{max}$ $\geq$ 60$^\circ$\\
        \hline
        4) Passage of cutting tool through the system & Working channel inner diameter ($D_3$) & $D_3$ $\geq$ 500$\mu$mm \\
        \hline
        5) Seamless workspace without dead zones & Decoupling degrees of freedom ($D_2$, $\alpha$) & $\mathbf{G}$($\alpha$ $\neq$ 0 $\Rightarrow$ $D_2$ $\geq$ 8mm) \\
        \hline
        6) Time-efficient task & Working channel tip angular velocity ($\dot{\alpha}$) & $\dot{\alpha}$ $\geq$ 10$^\circ$/s ($\raisebox{0.25ex}{\scalebox{0.575}{$\approx$}}$4mm/s)\\
        \hline
        7) Maintain position during tool insertion & Precision closed-loop control of $\alpha$ & $\left|\alpha(t) - \alpha_{c}\right|$ $\leq$ 2$^\circ$ ($\raisebox{0.25ex}{\scalebox{0.575}{$\approx$}}$1mm) \\
        \Xhline{1.6pt}
    \end{tabular}
    \vspace{-1mm}
\end{table}

\subsection{Balloon Design}

\begin{figure}[t!]
    \centering
    \includegraphics[width=\linewidth]{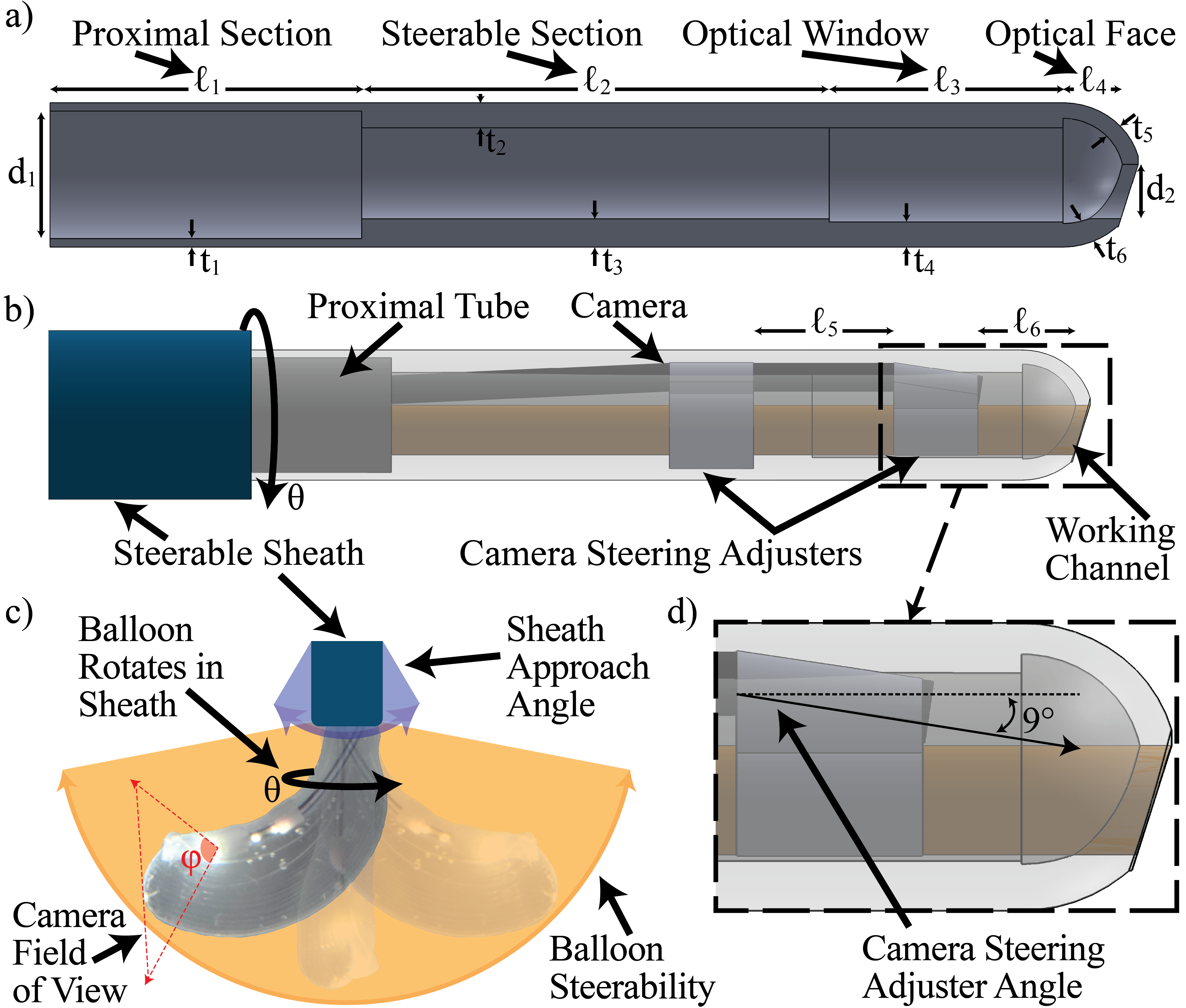}
    \vspace{-6mm}
    \caption{Balloon design and parameter definitions. (a) Cross section schematic showing design parameters. (b) Cardioscope schematic showing camera, working channel, and proximal tubing inserted through a steerable sheath. (c) Trumpet-shaped cardioscope workspace achieved via combined steering and rotation about its axis. (d)~Section view highlighting steering adjuster clip angle for camera alignment with optical face.}
    \vspace{0mm} 
    \label{FIGURE 2}
\end{figure}

\noindent
We designed a balloon using the schematic of Fig.~\ref{FIGURE 2}a and the parameters of Table~\ref{Table 2}. The collapsed (uninflated) outer diameter is 4.63mm, which meets our first design requirement of $D_{1}$ $\leq$ 5mm, indicating that it can fit through standard steerable sheaths as small as 15Fr inner diameter. The proximal end of the balloon is attached to a 0.13mm thick polymer (Pebax 5533) tube with a 3.8mm inner diameter that enables independent extension and rotation through the steerable sheath (Fig.~\ref{FIGURE 2}b). For optical clarity, the balloon endoscope is cast with Ecoflex\textsuperscript{TM}~00-45 Near Clear\textsuperscript{TM}. This material has ideal elastic properties and low mixed viscosity, which simplified parameterization to achieve the design metrics and ease of manufacturing, respectively. 
The parameters labeled in Fig.~\ref{FIGURE 2}a and quantified in Table~\ref{Table 2} were selected to decouple the optical face diameter expansion and working channel tip deflection angle by promoting expansion of the balloon in a prespecified sequence that is dependent on the thicknesses and section lengths (thinner sections of the balloon expand first). 

We tuned these parameters to meet design requirements (2-3 and 5), ensuring the target deployed face diameter, bending angle, and decoupling behavior (both avoiding a dead zone at low pressures and over-expansion at high pressures). Thickness t$_1$ was minimized to allow a maximal proximal tube size, maximizing internal space for the camera and working channel as defined by t$_2$ and t$_3$. 
Parameters t$_{4-6}$ and $\ell_{1-3}$ help enable the optical face to expand before steering occurs toward the thicker t$_3$ side at higher pressures.

The ballon system contains a chip camera for visualization, 3D printed clips (camera steering adjusters), and a working channel, which were all considered when tuning the parameters of the robot because each contributes some stiffness to the system. The 1mm diameter camera (Myriad Fiber Imaging endoscope with an OmniVision OV6946 integrated) has a 400x400 pixel resolution, 8-bit depth, and an integrated circumferential LED for illumination. The steering adjusters enable the optics to follow the tip deflection of the working channel and their dimensions were chosen to minimize their profile in the balloon (not shown to scale in Fig.~\ref{FIGURE 2}(b,d), printer resolution required slightly larger parts). 
The adjuster spacing (Fig.~\ref{FIGURE 2}b $\ell_{5-6}$) and view angle (Fig.~\ref{FIGURE 2}d) ensure the camera points toward the face of the balloon.

The working channel for tool passage was selected as a 1mm inner diameter, 0.5mm wall thickness transparent silicone tubing. When inflated and rotated around its axis by angle $\theta$, the system produces a trumpet-shaped workspace, as shown in Fig.~\ref{FIGURE 2}c, increasing steerability beyond what the sheath alone provides.

\begin{table}[t!]
    \centering
    \caption{Geometric values of tuned balloon (all dimensions in [$mm$]).}
    \label{Table 2}
    \vspace{-3mm}
    \begin{tabular}{!{\vrule width 1.6pt}C{0.16\linewidth}!{\vrule width 1.6pt}C{0.505\linewidth}!{\vrule width 1.6pt}C{0.16\linewidth}!{\vrule width 1.6pt}}
        \Xhline{1.6pt}
        Tunable Parameters    &     Parameter Descriptions (see~Fig.~\ref{FIGURE 2}(a-b)) &   Balloon Dimensions    \\ \Xhline{1.6pt}
                    $t_1$     &     Proximal Section Thickness              &     0.27                      \\ \hline
                    $t_2$     &     Top Steerable Section Thickness         &     0.80                      \\ \hline
                    $t_3$     &     Bottom Steerable Section Thickness      &     0.90                      \\ \hline
                    $t_4$     &     Bottom Optical Window Thickness         &     0.80                      \\ \hline
                    $t_5$     &     Top Optical Face Thickness              &     0.50                      \\ \hline
                    $t_6$     &     Bottom Optical Face Thickness           &     0.75                      \\ \hline
                    $d_1$     &     Proximal Section Inner Diameter         &     4.09                      \\ \hline
                    $d_2$     &     Distal Neck Inner Diameter              &     1.75                      \\ \hline
                    $\ell_1$  &     Proximal Section Length                 &     10.0                      \\ \hline  
                    $\ell_2$  &     Steerable Section Length                &     15.0                      \\ \hline
                    $\ell_3$  &     Optical Window Length                   &     7.50                      \\ \hline
                    $\ell_4$  &     Optical Face Length                     &     1.90                      \\ \hline
                    $\ell_5$  &     Straight Clip Distance to Angled Clip    &     5.00                      \\ \hline
                    $\ell_6$  &     Angled Clip Distance to Optical Face    &     4.00                      \\ \Xhline{1.6pt}
    \end{tabular}
    \vspace{-5mm}
\end{table}

\subsection{Cardioscope Fabrication} 

\begin{figure*}[t!]
    \centering
    \includegraphics[width=\textwidth]{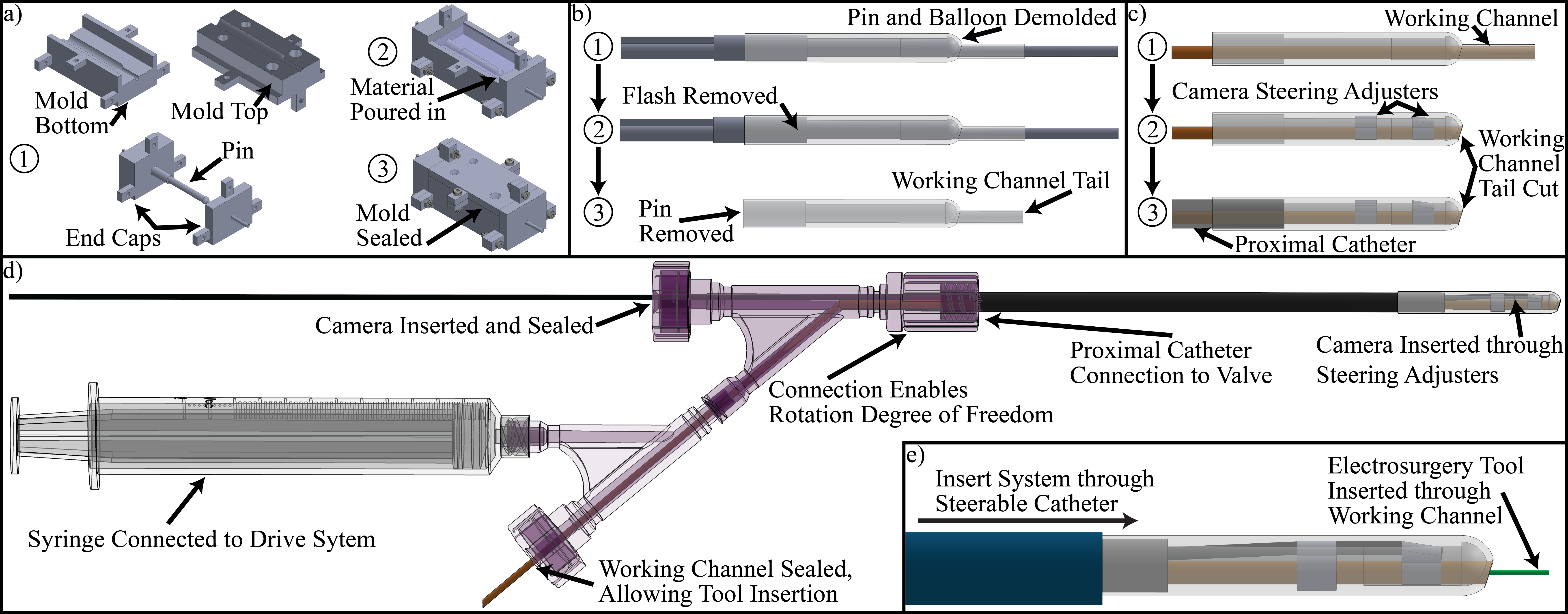}
    \vspace{-7mm}
    \caption{Cardioscope fabrication. (a)~Balloon casting. (b)~Ballon demolding. (c)~Post-processing and bonding of tubing connections. (d)~Assembly of complete cardioscope. (e)~Insertion through a steerable sheath.}
    \vspace{-2mm} 
    \label{FIGURE 3}
\end{figure*}

\noindent
To manufacture the balloon, we utilized a 3D printed mold consisting of five components, the top and bottom negatives (exterior) of the balloon and two end caps that align a pin within the top and bottom to create the positive (interior) of the balloon, as shown in Fig.~\ref{FIGURE 3}a$_1$. 
The pin is placed between the end caps with the proximal end of the balloon keyed by an internal feature within the cap to ensure longitudinal alignment within the negatives. 

The balloon material is mixed, degassed, and poured into the bottom of the mold and then the pin is placed into the material and the end caps are bolted to the mold bottom (Fig.~\ref{FIGURE 3}a$_2$). Then, as shown in Fig.~\ref{FIGURE 3}a$_3$, the mold top is placed and bolted to the bottom and end caps for the material to be left to cure for 4hrs at room temperature. After curing, the pin and balloon are demolded, flash is cut off the balloon to avoid adding undesired diameter and the pin is removed (Fig.~\ref{FIGURE 3}b). 
 
To assemble the complete robotic cardioscope, first the working channel is bonded to the balloon with SilPoxy\textsuperscript{TM}, then the steering adjusters are placed and the distal end of the working channel is cut flush with the optical face of the balloon (Fig.~\ref{FIGURE 3}c$_{1-2}$). Then, the proximal tube is bonded to the balloon with Gorilla Super Glue Gel (Fig.~\ref{FIGURE 3}c$_3$). Next, the assembled balloon proximal catheter is connected to a hemostasis valve Y connector luer lock that enables rotation of the system. The working channel extends through this connector and back out another hemostasis valve Y connector (with a male luer slip and female luer lock connecting to a syringe) and is sealed by a tuohy borst connection. 

The camera is then inserted in the balloon, placed into the steering adjusters, and proximally sealed via the tuohy borst, allowing for saline pressurization from volume infusion with the syringe (Fig.~\ref{FIGURE 3}d). 
Saline is chosen as the working fluid because it improves optical clarity through the balloon and is safe if released into the bloodstream. Lastly, the assembly can be passed through a steerable catheter and tools can be passed through the endoscope working channel, as shown in Fig.~\ref{FIGURE 3}e.

\section{Closed-loop Control}

\noindent
To enable closed-loop control of steering angle, a computer-controlled syringe pump was constructed as shown in Fig. \ref{FIGURE 4}a. The system is composed of 3D printed frame components holding the syringe with a linear guide rail and a threaded rod to produce incremental volume steps for characterization and calibration experiments or continuous flow for control and validation tests. 

To meet design requirement (6) for our targeted application, from Table~\ref{Table 1}, the system should bend from 0$^{\circ}$ to 60$^{\circ}$ in $\leq$ 6s for time-efficient task execution. To achieve this, we selected a Nema 17 stepper motor, which delivers up to 42Ncm of torque across speeds from 0-450rpm. This corresponds to flow rates from 0-3mL/s with our syringe and lead screw. 

An EASON Stepper Motor Driver was interfaced with Arduino and used to divide the 1.8$^\circ$ step angle the motor provides into 32 subdivisions, allowing for high-resolution volume delivery. The system was calibrated by verifying accurate saline delivery across infuse speeds to ensure consistent and repeatable syringe-controlled volume delivery prior to balloon endoscope integration. 

\subsection{Image-based Sensing}

\begin{figure}[tb!]
    \vspace{2mm}
    \centering
    \includegraphics[width=\linewidth]{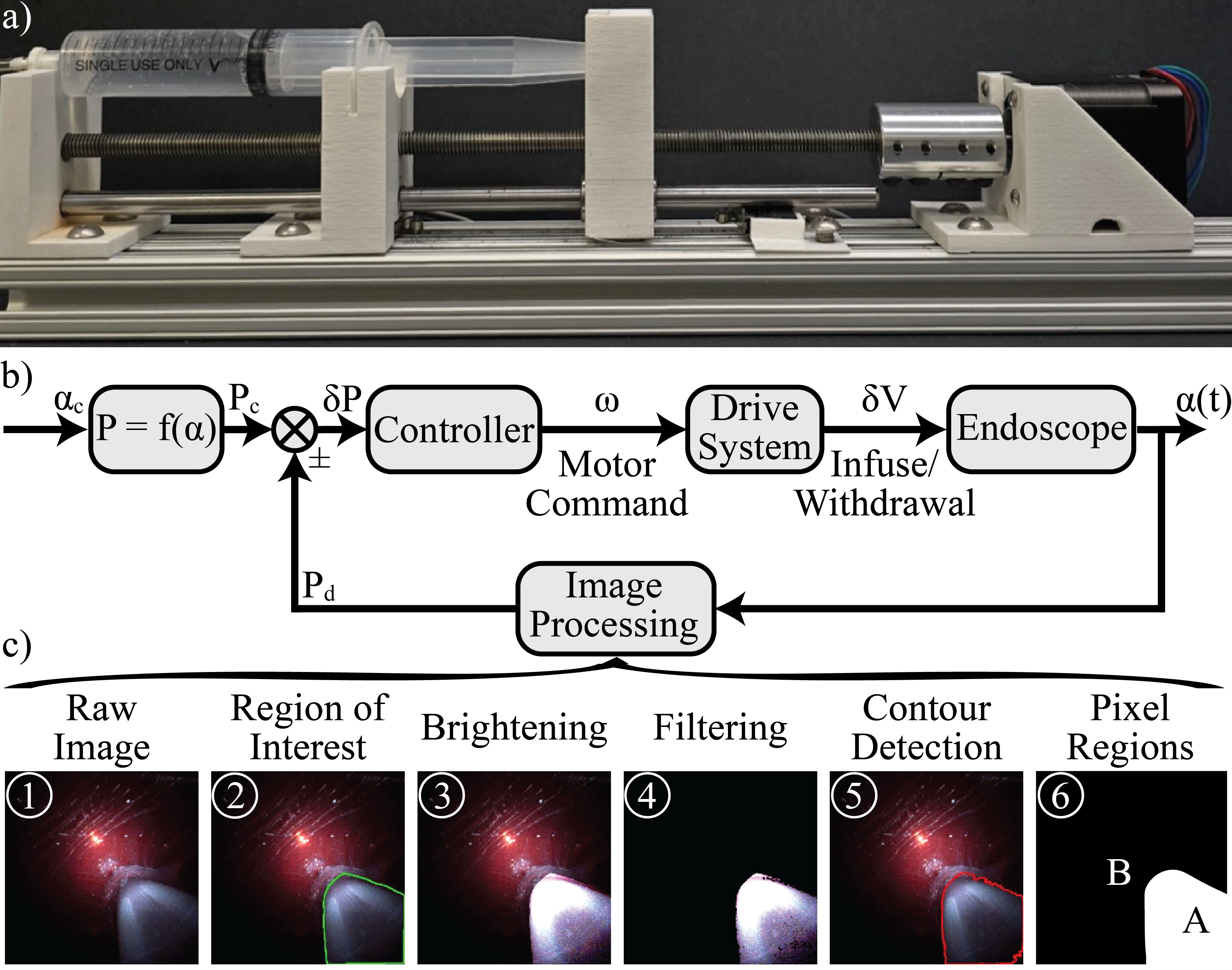}
    \vspace{-7mm}
    \caption{Steering angle control. (a) Computer-controlled infusion pump. (b) Controller block diagram. (c) Image processing pipeline for estimating steering angle.}
    \vspace{-5mm} 
    \label{FIGURE 4}
\end{figure}

\noindent
Owing to their bending stiffness, tools inserted and removed from the working channel can cause the cardioscope to deflect. To enable automatic steering compensation during tool insertion and removal, an image-based sensing approach was developed using the embedded camera to track slight steering-angle-dependent motions of the working channel in the image frame. Fig.~\ref{FIGURE 4}b shows the closed-loop control feedback loop where, given a commanded tip deflection angle ($\alpha_c$) the controller utilizes the image processing steps in Fig.~\ref{FIGURE 4}c to steer the actual angle ($\alpha(t)$) 
to the commanded orientation. 

The steps of the image processing algorithm are as follows. First, a region of interest around the working channel (defined during calibration and shown as a green line in Fig.~\ref{FIGURE 4}c$_{2}$) is segmented and each pixel value is brightened by a factor of 3.5 (capped at 255), see Fig.~\ref{FIGURE 4}c$_{3}$. 
Next, the pixels are filtered using three different approaches to convert any pixels that do not correspond to the working channel to black (removed by setting their pixel value to 0). First, red pixels (corresponding to blood) were detected in HSV color space using hue thresholds $\leq$10 or $\geq$160, and saturation $>$15. 
Second, near-black pixels, corresponding to the dark background at the camera edge, were removed based on intensity, defined as grayscale values $<$5. Third, pixels outside the region of interest were removed. The filtered image is shown in Fig.~\ref{FIGURE 4}c$_{4}$. 
Finally, as shown in Fig.~\ref{FIGURE 4}c$_5$, a contour region is drawn around the working channel using OpenCV's \verb|findContours| function applied to a grayscale version of the image. 
The number of pixels inside, $P_A$, and outside, $P_B$, the working channel (Fig.~\ref{FIGURE 4}c$_6$) are used to compute the pixel ratio ($P$) as 
\begin{equation}
    P = \Big(\frac{P_A}{P_{Total}}\Big) = \Big(\frac{P_A}{P_A + P_B}\Big)
    \label{EQ 1}
\end{equation}
in which $P_{Total}=160k$ pixels. As will be shown in the section below, pixel ratio is functionally related to steering angle, $\alpha$. 
\begin{equation}
    P = f(\alpha).
    \label{EQ 2}
\end{equation}
The control system of Fig.~\ref{FIGURE 4}b uses this function to estimate the angular error and to apply a motor command that adjusts the infused volume to reduce the error. Controller details are presented in the next section. 

\section{Experiments}

\noindent
To characterize the steerable cardioscope, a sequence of experiments were performed. The first set of experiments evaluated the success of the design in using a single input, inflation volume or pressure, to independently control optical diameter and steering angle. The second set of experiments involved calibrating, implementing and assessing closed-loop control of steering angle. A final set of experiments considered the quality of the cardioscopic images in the context of aortic leaflet visualization. Each set of experiments is described in the subsections below.

\begin{figure}[tb!]
    \centering    \includegraphics[width=\linewidth]{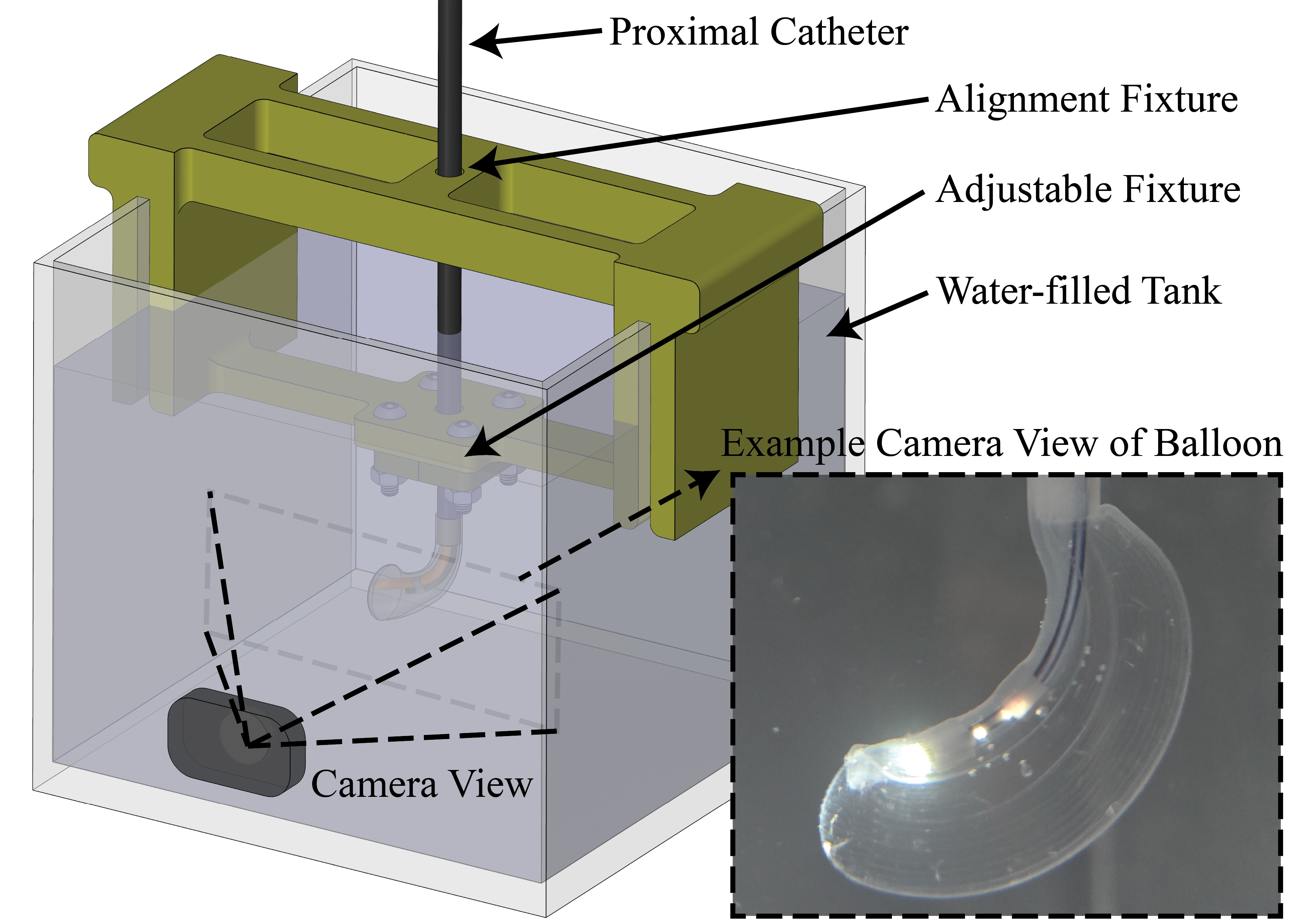}
    \vspace{-6mm}
    \caption{Experimental platform. Cardioscope is suspended in saline tank with external camera used to record balloon size and steering angle.}
    \vspace{-4mm} 
    \label{FIGURE X}
\end{figure}

As shown in the schematic of Fig. \ref{FIGURE X}, experiments were conducted in an acrylic tank filled with water, allowing visual monitoring of system performance with an external camera. For experiments requiring measurement of balloon dimensions and steering angle, a manual vector-based image analysis process (based on~\cite{mccandless_soft_2022}) was used on recorded images and videos to extract ground truth measurements of tip deflection angle and optical window face diameter. A Savitzky-Golay filter with a 2\textsuperscript{nd} order polynomial and a window size of seven was applied to reduce noise while preserving signal shape for visualization of the results.

\vspace{-1mm}
\subsection{Assessing Independent Control of Optical Window Face Diameter and Steering Angle}

\begin{figure}[tb!]
    \centering
    \includegraphics[width=\linewidth]{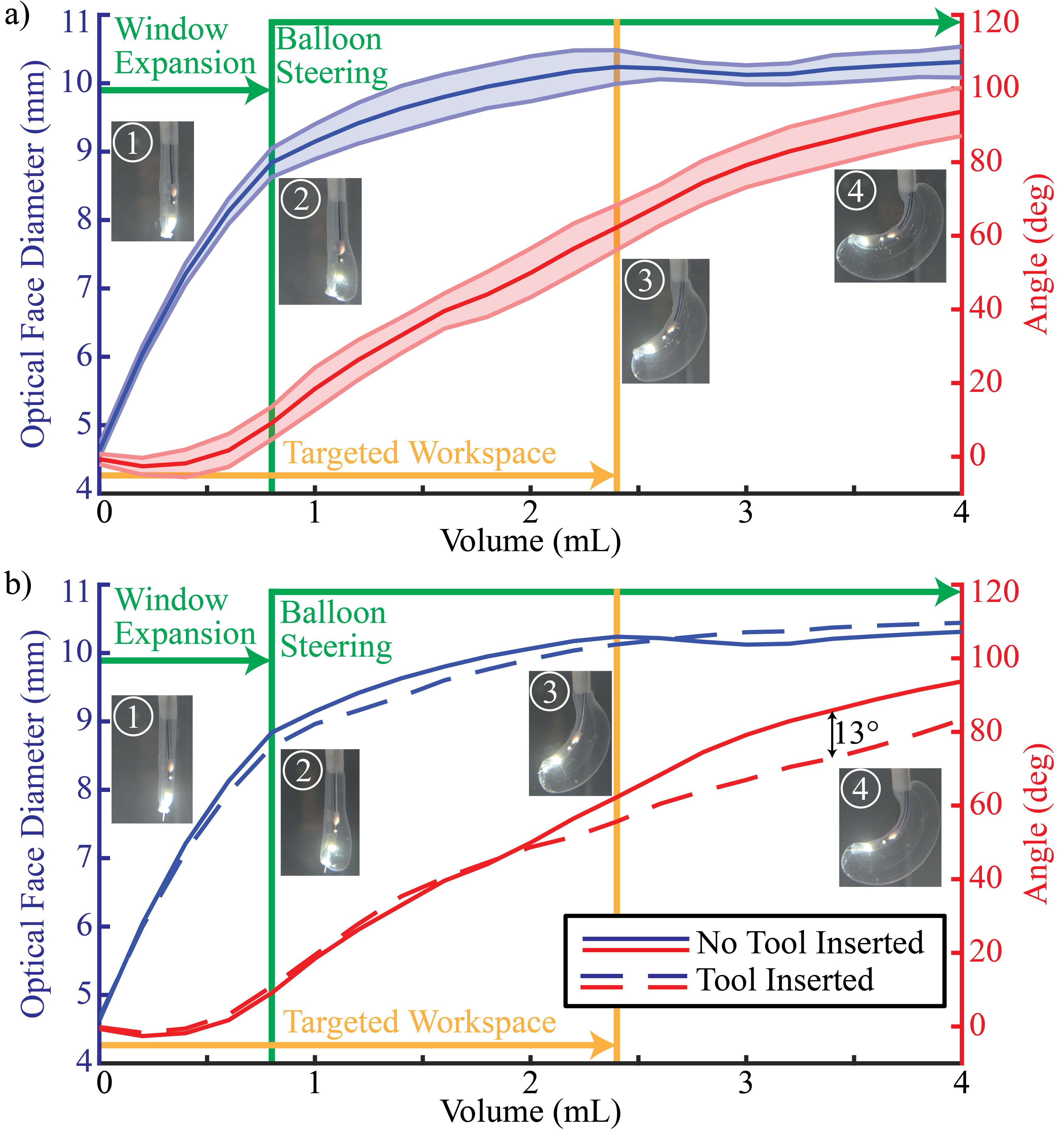}
    \vspace{-7mm}
    \caption{Optical window face diameter and tip deflection angle as functions of inflation volume. (a)~Cardioscope without a tool inserted. The solid line is the mean value and the shaded area is the standard deviation. (b)~Cardioscope with a tool (electrosurgical wire) inserted.}
    \vspace{2mm} 
    \label{FIGURE 5}
\end{figure}

\noindent
Ten trials were performed in which balloon infusion volume was varied from 0 to 4ml in increments of 0.2ml. For each infusion volume, balloon images were collected using the external camera. Experiments were performed with and without a tool (360 $\mu$m electrosurgical wire) inserted through the working channel. Fig.~\ref{FIGURE 5} depicts the results showing optical face diameter and working channel tip deflection angle as functions of inflation volume.

With increasing volume, initially the optical face expands from 0-0.8mL (Fig.~\ref{FIGURE 5}a$_{1-2}$), then balloon deflection occurs for higher inflation volumes up to $\approx$100$^\circ$ at 4mL (Fig.~\ref{FIGURE 5}a$_4$), while optical face diameter levels off over that range. 
The successful decoupling of the degrees of freedom (meeting design requirement (5)) with a single volume input is highlighted by the vertical green lines in the plot, which also show that $D_2$~$\geq$~8mm before $\alpha \geq$~0 and that 8mm $\leq$ $D_2$ $\leq$ 11mm (meeting design requirement (2)). 
Additionally, we see that a bending angle $\alpha \geq$~60$^\circ$ is achieved (Fig.~\ref{FIGURE 5}a$_3$), as shown by the vertical orange line indicating this targeted workspace, meaning that design requirement (3) was met.

Fig.~\ref{FIGURE 5}b depicts the system with an electrosurgical wire (cutting tool) passed through the 1mm diameter working channel (see inset images Fig.~\ref{FIGURE 5}b$_{1-4}$). The main difference from Fig.~\ref{FIGURE 5}a, as highlighted in Fig.~\ref{FIGURE 5}b$_4$, is a maximum 13$^\circ$ difference in steering angle for the same inflation volume. This difference is substantial enough to indicate that closed-loop control of bending angle will be beneficial. At the same time, it is not so large that the design requirements of Table I are not satisfied: Increasing the inflation volume to compensate for the tool will only slightly increase the optical face diameter.

\subsection{Closed-loop Control of Steering Angle}
\vspace{0mm}
    
\begin{figure}[tb!]
    \centering
    \includegraphics[width=\linewidth]{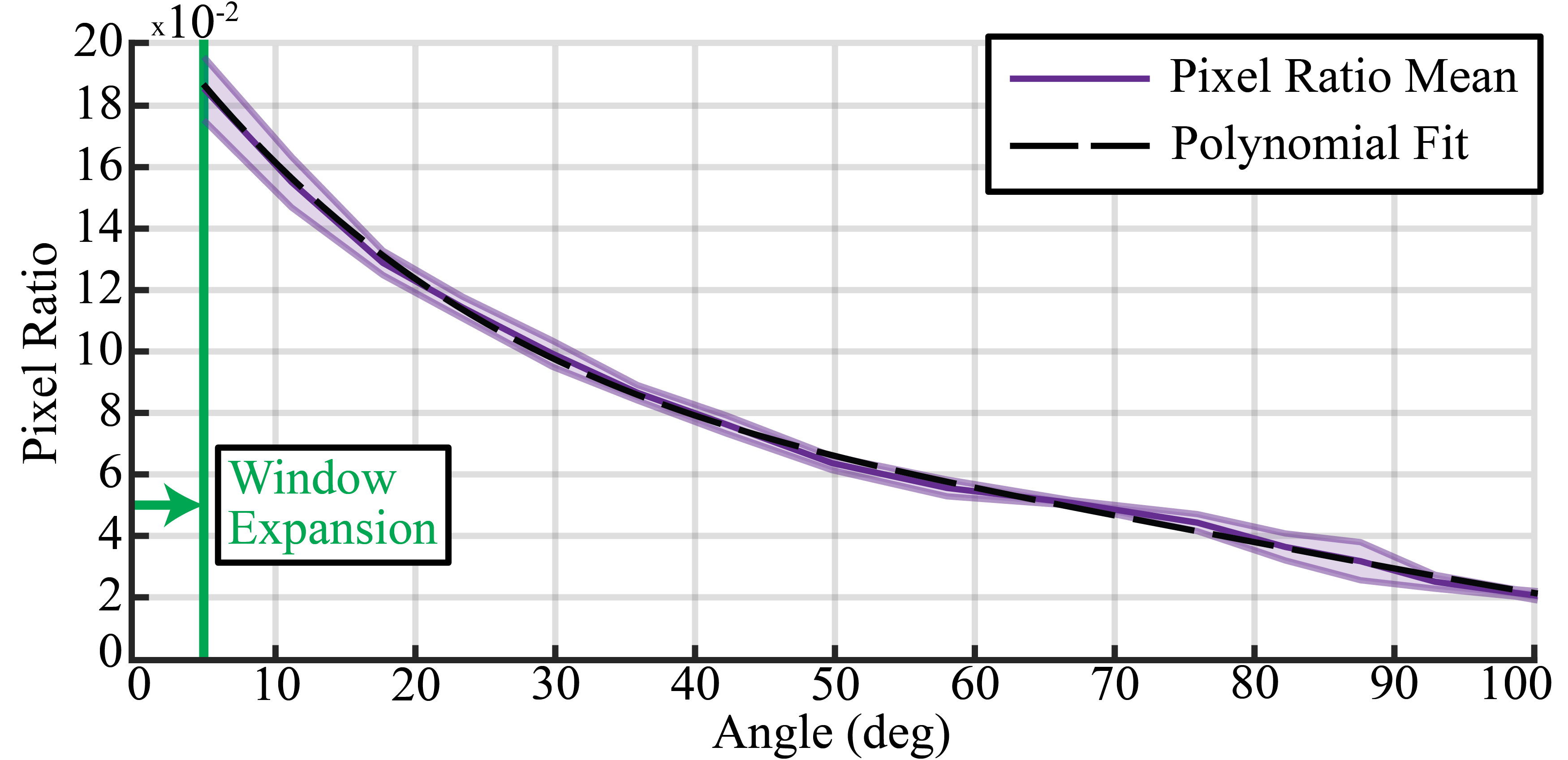}
    \vspace{-7mm}
    \caption{Pixel ratio as a function of steering angle. Mean and standard deviation for 3 trials are depicted along with a 4th order polynomial fit.}
    \vspace{0mm} 
    \label{FIGURE 6}
\end{figure}

\noindent
To implement the controller of Fig. \ref{FIGURE 4}, the relationship between pixel ratio and bending angle, given by (\ref{EQ 1}) and (\ref{EQ 2}), is needed. To obtain this relationship under conditions representative of the blood-filled heart, the water in the testing tank was dyed red (FD\&C Red 40, Ward’s Science) while still allowing visual recording using the external camera. Pixel ratio as a function of bending angle was collected and the results are shown in Fig.~\ref{FIGURE 6} along with a 4$^{th}$ order polynomial fit. We note that, while the calibration curve varied slightly between different balloons, the curve is the same with and without a tool inserted in the working channel. 

Given this calibration function, the remaining part of the controller to define is the block relating the error in pixel ratio, $\delta P$, to the commanded motor velocity, $\omega$, given in rpm. The following multi-threshold bang-bang control law was tuned to meet the design requirement of steering the balloon 60$^\circ$ in less than 1s while remaining stable and not producing overshoot. 

\vspace{0mm}
\begin{equation}
    \omega = 
        \begin{cases} 
        100 \sgn(\delta P) & \text{if } |\delta P| > .006 \\
        25 \sgn(\delta P)  & \text{if } .002 \leq |\delta P| \leq .006 \\
        5  \sgn(\delta P)  & \text{if } .001 \leq |\delta P| \leq .002 \\
        0  \sgn(\delta P)  & \text{if } |\delta P| < .001 \\
        \end{cases}
    \label{EQ 3}
\end{equation}

\begin{figure}[tb!]
    \centering
    \includegraphics[width=\linewidth]{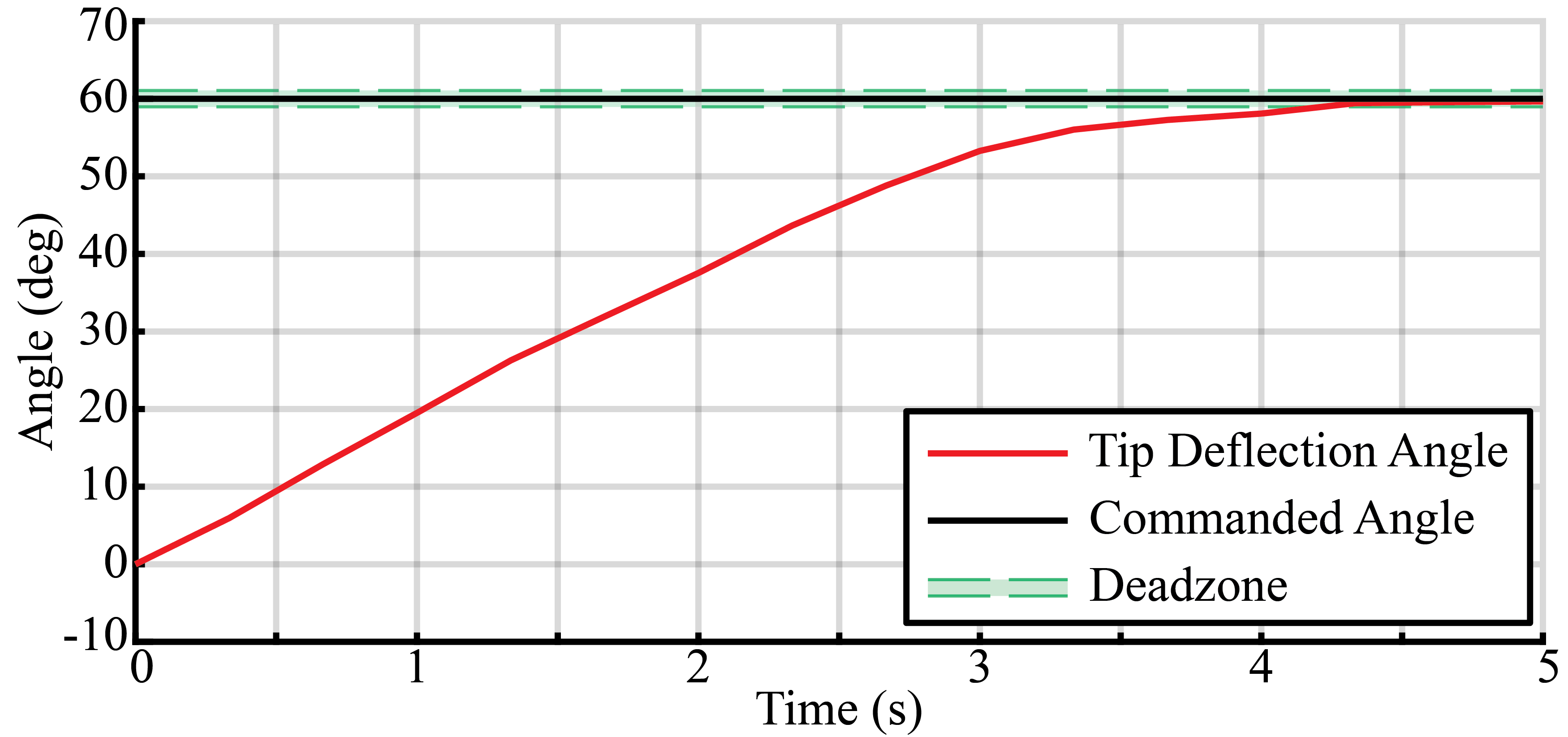}
    \vspace{-7mm}
    \caption{Step response for $60^\circ$ change in steering angle averaged over five trials.}
    \vspace{0mm} 
    \label{FIGURE 8}
\end{figure}

To validate performance of the control law, three experiments were undertaken. To ensure that the system met the steering speed requirement of $\dot{\alpha}\geq 10^\circ$/s, five trials were performed in which the steering angle was initially at zero and a step input of 60$^\circ$ was commanded. The response is depicted in Fig.~\ref{FIGURE 8} with steering angle based on video analysis from the external camera. With an overdamped response settling to 60$^\circ$ in under 6s (average tip velocity =  $\approx$14$^\circ$/s), the system meets design requirement 6 of Table I.

\begin{figure}[tb!]
    \centering
    \includegraphics[width=\linewidth]{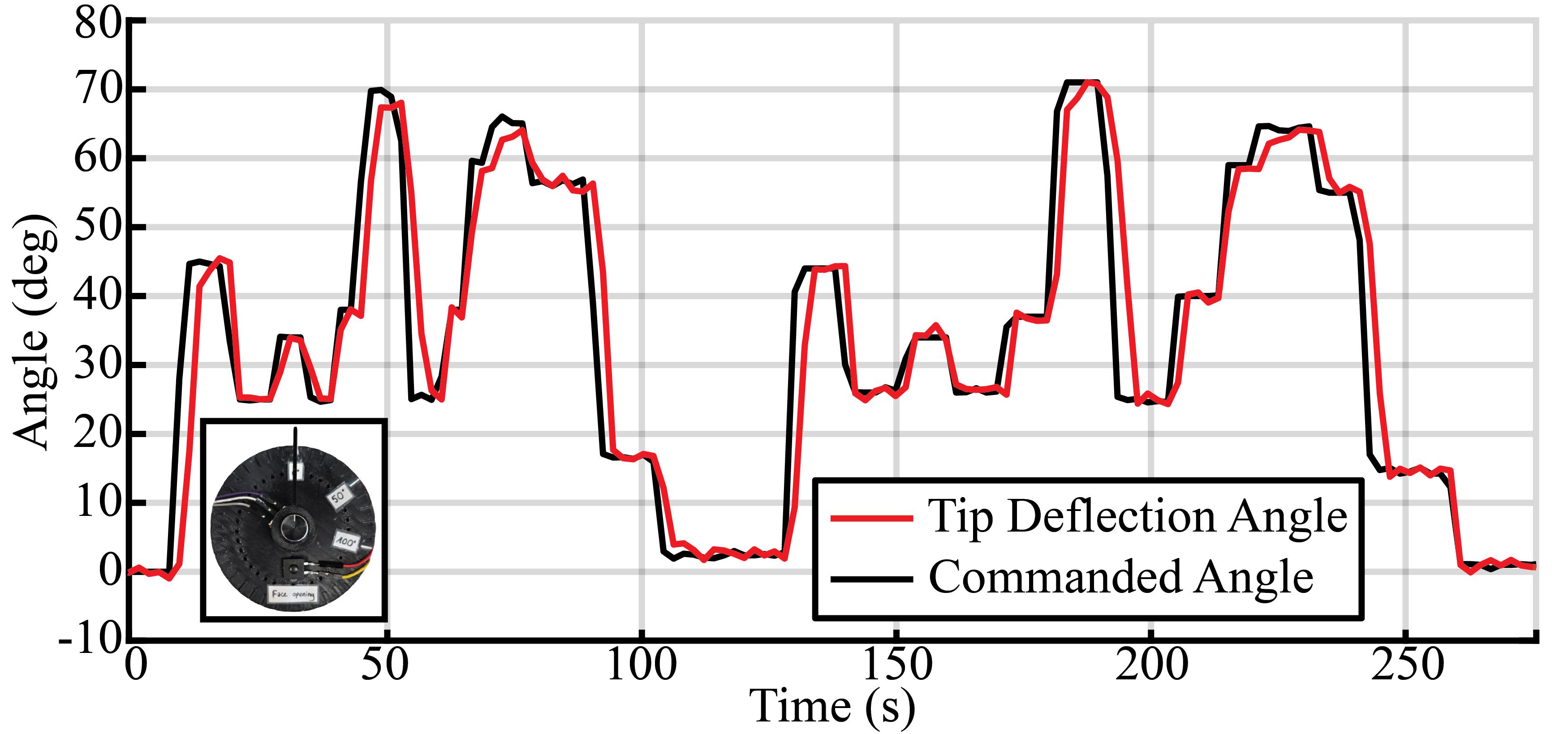}
    \vspace{-7mm}
    \caption{Response to operator commanded step changes in steering angle using the rotary knob shown in the inset.}
    \vspace{0mm} 
    \label{FIGURE 11}
\end{figure}

To understand how the system would respond to multiple smaller step commands, an experiment was performed in which the operator used a control interface in which pressing a button caused the optical window to inflate and a rotary knob allowed adjustment of steering angle between 0-100$^\circ$. Fig.~\ref{FIGURE 11} plots the commanded and measured steering angles, as computed using pixel ratio. It is observed that the system responds quickly to both small and large requested changes in steering angle. 

\begin{figure}[tb!]
    \centering
    \includegraphics[width=\linewidth]{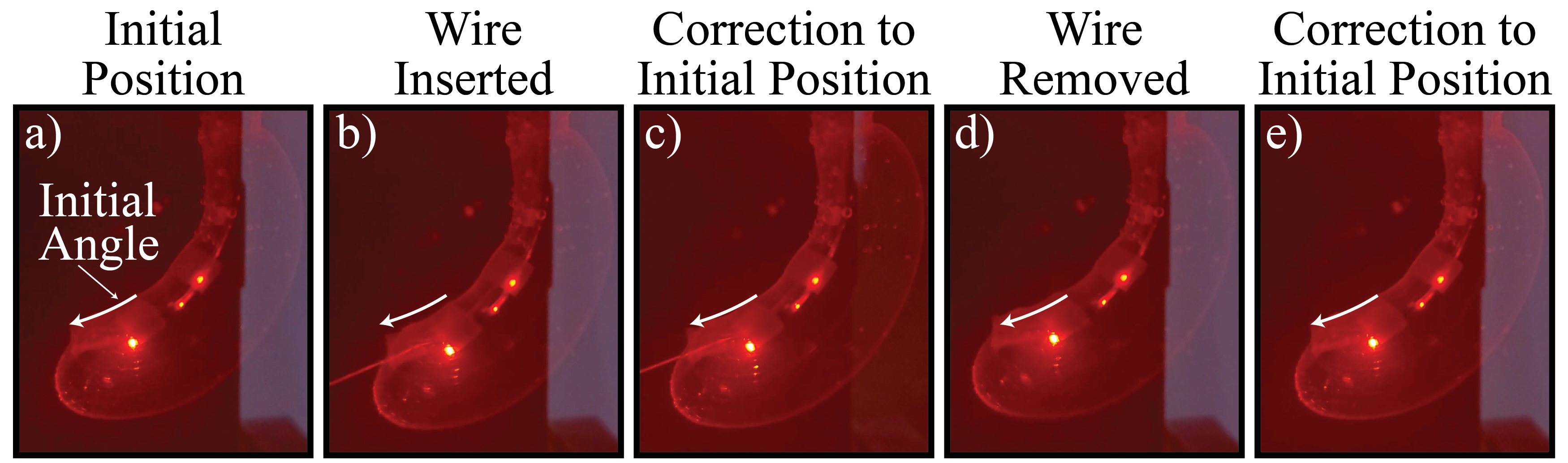}
    \vspace{-7mm}
    \caption{Automatic compensation of balloon deflection to maintain orientation during tool insertion and removal. (a) Initial steering angle with working channel edge labeled with a white arrow (label repeated in subsequent images). 
    (b) Balloon straightens due to wire insertion. (c) Controller infuses balloon to correct steering angle. (d) Balloon increases in steering angle when wire is removed. (e) Controller reduces balloon infusion to bring steering angle back to desired value.}
    \vspace{-4mm} 
    \label{FIGURE 9}
\end{figure}

Finally, Fig.~\ref{FIGURE 9} shows the ability of the controller to automatically compensate for the change in steering angle that occurs with tool insertion and removal. Tool insertion causes the balloon to straighten due to the bending stiffness of the tool. The controller detects this straightening as a change in pixel ratio and reacts by increasing the infused volume. When the tool is removed, the increase in infusion volume to accommodate the tool causes the balloon steering angle to exceed the desired value. The controller detects this oversteering and reacts to reduce the infused volume returning the steering angle to desired value.  
From the figure, it can be observed that design requirement 7 of Table I is met: tool induced steering angle error is less than $2^\circ$.

\subsection{Cardioscope Image Quality}

\begin{figure}[tb!]
    \centering    
    \includegraphics[width=\linewidth]{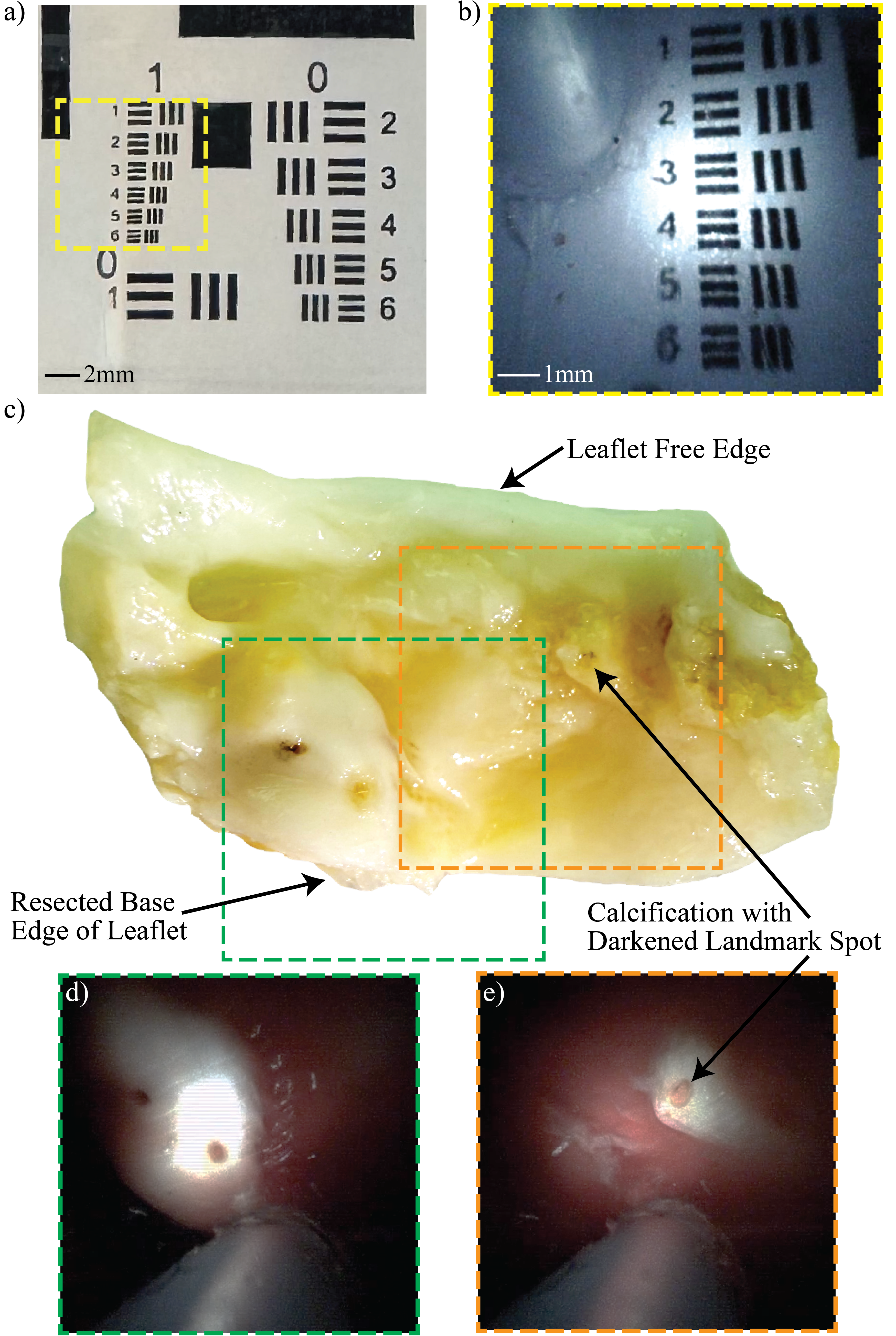}
    \vspace{-7.5mm}
    \caption{Cardioscopic visualization. (a) Laminated USAF target placed within water-filled beaker. (b) Cardioscopic image of the target. (c) Calcified human aortic leaflet. (d-e) Corresponding cardioscopic images of the leaflet when fully submerged in blood. Calcifications include darkened spots where traversal was attempted.}
    \vspace{-3mm} 
    \label{FIGURE 10}
\end{figure}

\noindent
As a measure of optical clarity, a paper 1951 USAF resolution test chart was laminated and placed in a water-filled beaker, as shown in Fig.~\ref{FIGURE 10}a. 
The cardioscope was then used to image the six elements of group 1 on the target (Fig.~\ref{FIGURE 10}b). The cardiscope can clearly resolve these features, the smallest of which are 0.14mm wide lines spaced 0.14mm apart. 

As shown in Fig.~1b, to test the capability of targeting tissue locations with the working channel, a laminataed paper bullseye pattern with a central ring diameter of 1mm and 0.25mm wide ring lines spaced 0.25mm was placed under the beaker and the cardioscope was manually adjusted to center the working channel over the bullseye. The cardioscopic image enables accurate positioning of the working channel with respect to a target (Fig.~1c). 

To assess image quality for the clinical application of leaflet laceration, a heavily calcified human leaflet obtained during surgical aortic valve replacement (Fig.~\ref{FIGURE 10}c) was submerged in porcine blood and a sequence of cardioscopic images was obtained (a subset of the images is shown in Fig.~\ref{FIGURE 10}d-e). 
The calcifications can be clearly visualized, allowing an operator to avoid them when attempting electrosurgical penetration of the leaflet. 

\section{CONCLUSIONS}

\noindent
This paper introduced a new class of cardioscopes that combine the benefits of vascular delivery, inclusion of a working channel, and steerability. The design leverages variations in balloon wall thickness to use a single input, inflation volume, to provide decoupled control of optical window diameter and steering angle. 

Incorporating steering into the balloon cardioscope enables robotic catheter designs combining tendon-actuated sheaths for tip positioning with cardioscopic balloons for controlling tip orientation. While the fabricated design was tailored to aortic leaflet laceration, the 3D printing-based mold fabrication process can be easily adapted to many other procedures through judicious selection of parameters \{$t_i$,$\ell_j$,$d_k$\}. In fact, the balloon presented is already capable of bending greater than 90$^\circ$ and so could be used for a variety of applications, such as those requiring coronary sinus access. 

As demonstrated in the visualization experiments, the direct optical imaging provided by the system can enable precise targeting inside the heart. While illustrated using a tool comprising an electrosurgical wire, the working channel could be used to deliver a wide variety of instruments such as optical fibers~\cite{mccandless_comparison_2025}, stents, occlusion devices and microcatheters. Feedback control of steering angle is image-based and so does not require an additional sensor. Furthermore, the controller is agnostic to the type of device delivered and so can facilitate high-precision tool positioning and easy integration with a robotic catheter system. 

\section*{ACKNOWLEDGMENTS}

\vspace{-1mm}
\noindent
Institutional Review Number: IRB-P00038048, initially approved on March 5, 2021, and reapproved on April 19, 2024.

Fig.~\ref{FIGURE 1}d utilizes a modified BioRender.com heart template.

\bibliographystyle{IEEEtran.bst}
\vspace{-1mm}
\typeout{}
\bibliography{IEEEabrv.bib, referencesCHILDRENS.bib}

\begin{thebibliography}{10}
\providecommand{\url}[1]{#1}
\csname url@rmstyle\endcsname
\providecommand{\newblock}{\relax}
\providecommand{\bibinfo}[2]{#2}
\providecommand\BIBentrySTDinterwordspacing{\spaceskip=0pt\relax}
\providecommand\BIBentryALTinterwordstretchfactor{4}
\providecommand\BIBentryALTinterwordspacing{\spaceskip=\fontdimen2\font plus
\BIBentryALTinterwordstretchfactor\fontdimen3\font minus \fontdimen4\font\relax}
\providecommand\BIBforeignlanguage[2]{{%
\expandafter\ifx\csname l@#1\endcsname\relax
\typeout{** WARNING: IEEEtran.bst: No hyphenation pattern has been}%
\typeout{** loaded for the language `#1'. Using the pattern for}%
\typeout{** the default language instead.}%
\else
\language=\csname l@#1\endcsname
\fi
#2}}

\bibitem{harky_future_2020}
A.~Harky, G.~Chaplin, J.~S.~K. Chan, P.~Eriksen, B.~MacCarthy-Ofosu, T.~Theologou, and A.~D. Muir, ``\BIBforeignlanguage{en}{The {Future} of {Open} {Heart} {Surgery} in the {Era} of {Robotic} and {Minimal} {Surgical} {Interventions}},'' \emph{\BIBforeignlanguage{en}{Heart, Lung and Circulation}}, vol.~29, no.~1, pp. 49--61, 2020.

\bibitem{nayar_ultrasound-guided_2024}
N.~U. Nayar and J.~P. Desai, ``Ultrasound-{Guided} {Real}-{Time} {Joint} {Space} {Control} of a {Robotic} {Transcatheter} {Delivery} {System},'' \emph{IEEE Robotics and Automation Letters}, vol.~9, no.~9, pp. 7677--7684, 2024.

\bibitem{chen_first--human_2024}
X.~Chen, M.~Su, X.~Chen, B.~Wang, M.~Taramasso, and Y.~Wang, ``\BIBforeignlanguage{en}{First-in-human robotic-assisted transcatheter mitral edge-to-edge repair for treatment of severe mitral regurgitation},'' \emph{\BIBforeignlanguage{en}{The International Journal of Cardiovascular Imaging}}, vol.~40, no.~12, pp. 2641--2644, 2024.

\bibitem{pittiglio_magnetic_2024}
G.~Pittiglio, F.~Leuenberger, M.~Mencattelli, M.~McCandless, E.~O’Leary, and P.~E. Dupont, ``Magnetic {Ball} {Chain} {Robots} for {Cardiac} {Arrhythmia} {Treatment},'' \emph{IEEE Transactions on Medical Robotics and Bionics}, vol.~6, no.~4, pp. 1322--1333, 2024.

\bibitem{rogatinsky_multifunctional_2023}
J.~Rogatinsky, D.~Recco, J.~Feichtmeier, Y.~Kang, N.~Kneier, P.~Hammer, E.~O’Leary, D.~Mah, D.~Hoganson, N.~V. Vasilyev, and T.~Ranzani, ``\BIBforeignlanguage{en}{A multifunctional soft robot for cardiac interventions},'' \emph{\BIBforeignlanguage{en}{Science Advances}}, vol.~9, no.~43, p. eadi5559, 2023.

\bibitem{li_robust_2025}
Z.~Li, C.~Lambranzi, D.~Wu, A.~Segato, F.~De~Marco, E.~V. Poorten, J.~Dankelman, and E.~De~Momi, ``Robust {Path} {Planning} via {Learning} {From} {Demonstrations} for {Robotic} {Catheters} in {Deformable} {Environments},'' \emph{IEEE Transactions on Biomedical Engineering}, vol.~72, no.~1, pp. 324--336, 2025.

\bibitem{young_robotic_2022}
L.~Young and J.~Khatri, ``\BIBforeignlanguage{en}{Robotic {Percutaneous} {Coronary} {Intervention}: {The} {Good}, the {Bad}, and {What} is to {Come}},'' \emph{\BIBforeignlanguage{en}{US Cardiology Review}}, vol.~16, p. e02, 2022.

\bibitem{vasilyev_three-dimensional_2006}
N.~V. Vasilyev, J.~F. Martinez, F.~P. Freudenthal, Y.~Suematsu, G.~R. Marx, and P.~J. Del~Nido, ``\BIBforeignlanguage{en}{Three-{Dimensional} {Echo} and {Videocardioscopy}-{Guided} {Atrial} {Septal} {Defect} {Closure}},'' \emph{\BIBforeignlanguage{en}{The Annals of Thoracic Surgery}}, vol.~82, no.~4, pp. 1322--1326, 2006.

\bibitem{sorensen_camera--tip_2025}
S.~T. Sørensen, W.~Messina, L.~Niemitz, C.~O’Dowling, P.~Buszman, S.~Andersson-Engels, and R.~Burke, ``\BIBforeignlanguage{en}{Camera-on-tip endoscope for \textit{in vivo} cardiovascular diagnostics and surgical guidance},'' \emph{\BIBforeignlanguage{en}{Biomedical Optics Express}}, vol.~16, no.~1, p.~12, 2025.

\bibitem{rosa_cardioscopically_2017}
B.~Rosa, Z.~Machaidze, M.~Mencattelli, S.~Manjila, B.~Shin, K.~Price, M.~A. Borger, V.~Thourani, P.~Del~Nido, D.~W. Brown, C.~W. Baird, J.~E. Mayer, and P.~E. Dupont, ``\BIBforeignlanguage{en}{Cardioscopically {Guided} {Beating} {Heart} {Surgery}: {Paravalvular} {Leak} {Repair}},'' \emph{\BIBforeignlanguage{en}{The Annals of Thoracic Surgery}}, vol. 104, no.~3, pp. 1074--1079, 2017.

\bibitem{machaidze_optically-guided_2019}
Z.~Machaidze, M.~Mencattelli, G.~Arnal, K.~Price, F.-Y. Wu, V.~Weixler, D.~W. Brown, J.~E. Mayer, and P.~E. Dupont, ``\BIBforeignlanguage{en}{Optically-guided instrument for transapical beating-heart delivery of artificial mitral chordae tendineae},'' \emph{\BIBforeignlanguage{en}{The Journal of Thoracic and Cardiovascular Surgery}}, vol. 158, no.~5, pp. 1332--1340, 2019.

\bibitem{fagogenis_autonomous_2019}
G.~Fagogenis, M.~Mencattelli, Z.~Machaidze, B.~Rosa, K.~Price, F.~Wu, V.~Weixler, M.~Saeed, J.~E. Mayer, and P.~E. Dupont, ``\BIBforeignlanguage{en}{Autonomous robotic intracardiac catheter navigation using haptic vision},'' \emph{\BIBforeignlanguage{en}{Science Robotics}}, vol.~4, no.~29, p. eaaw1977, 2019.

\bibitem{padala_transapical_2012}
M.~Padala, J.~H. Jimenez, A.~P. Yoganathan, A.~Chin, and V.~H. Thourani, ``\BIBforeignlanguage{en}{Transapical beating heart cardioscopy technique for off-pump visualization of heart valves},'' \emph{\BIBforeignlanguage{en}{The Journal of Thoracic and Cardiovascular Surgery}}, vol. 144, no.~1, pp. 231--234, 2012.

\bibitem{uchida_recent_2011}
Y.~Uchida, ``\BIBforeignlanguage{en}{Recent {Advances} in {Percutaneous} {Cardioscopy}},'' \emph{\BIBforeignlanguage{en}{Current Cardiovascular Imaging Reports}}, vol.~4, no.~4, pp. 317--327, 2011.

\bibitem{betensky_use_2012}
B.~P. Betensky, M.~Jauregui, B.~Campos, J.~Michele, F.~E. Marchlinski, L.~Oley, B.~Wylie, D.~Robinson, and E.~P. Gerstenfeld, ``\BIBforeignlanguage{en}{Use of a {Novel} {Endoscopic} {Catheter} for {Direct} {Visualization} and {Ablation} in an {Ovine} {Model} of {Chronic} {Myocardial} {Infarction}},'' \emph{\BIBforeignlanguage{en}{Circulation}}, vol. 126, no.~17, pp. 2065--2072, 2012.

\bibitem{khan_transcatheter_2018}
J.~M. Khan, D.~Dvir, A.~B. Greenbaum, V.~C. Babaliaros, T.~Rogers, G.~Aldea, M.~Reisman, G.~B. Mackensen, M.~H. Eng, G.~Paone, D.~D. Wang, R.~A. Guyton, C.~M. Devireddy, W.~H. Schenke, and R.~J. Lederman, ``\BIBforeignlanguage{en}{Transcatheter {Laceration} of {Aortic} {Leaflets} to {Prevent} {Coronary} {Obstruction} {During} {Transcatheter} {Aortic} {Valve} {Replacement}: {Concept} to {First}-in-{Human}},'' \emph{\BIBforeignlanguage{en}{JACC: Cardiovascular Interventions}}, vol.~11, no.~7, pp. 677--689, 2018.

\bibitem{mccandless_soft_2022}
M.~McCandless, A.~Perry, N.~DiFilippo, A.~Carroll, E.~Billatos, and S.~Russo, ``\BIBforeignlanguage{en}{A {Soft} {Robot} for {Peripheral} {Lung} {Cancer} {Diagnosis} and {Therapy}},'' \emph{\BIBforeignlanguage{en}{Soft Robotics}}, vol.~9, no.~4, pp. 754--766, 2022.

\bibitem{mccandless_comparison_2025}
M.~McCandless, S.~Otto, S.~Elmariah, N.~Langer, and P.~E. Dupont, ``A comparison of laser, electrosurgical, and mechanical transcatheter traversal of calcified aortic valve leaflets,'' \emph{JTCVS Structural and Endovascular}, vol.~6, p. 100054, 2025.

\end{thebibliography}

\end{document}